\documentclass[review,number,sort&compress,12pt]{elsarticle}

\usepackage{natbib}
\usepackage{geometry}
\usepackage{graphicx}
\usepackage{subfigure}
\usepackage{scrtime}
\usepackage{epstopdf}
\usepackage{enumerate}
\usepackage{amsmath}
\usepackage{siunitx}
\usepackage{amssymb}
\usepackage{upgreek}
\usepackage{appendix}
\usepackage[colorlinks, linkcolor=black, citecolor=black, anchorcolor=black]{hyperref}
\hypersetup{hidelinks}
\usepackage{algorithm}% http://ctan.org/pkg/algorithms
\usepackage{algorithmicx}
\usepackage{algpseudocode}% http://ctan.org/pkg/algorithmicx

\usepackage{multirow}
\usepackage{lineno}
\usepackage{enumerate}
\usepackage[section]{placeins}

\usepackage{booktabs}
\usepackage{threeparttable}

\usepackage{booktabs}
\usepackage[dvipsnames]{xcolor}
\usepackage{tcolorbox}
\usepackage{soul}
\usepackage{makecell}
\usepackage{mathtools}
\def\romannumer#1{\uppercase\expandafter{\romannumeral#1}}

\makeatletter
\def\ps@pprintTitle{
	\let\@oddhead\@empty
	\let\@evenhead\@empty
	\def\@oddfoot{\centerline{\thepage}}%
	\let\@evenfoot\@oddfoot}
\makeatother

\begin{document}
% \modulolinenumbers[1]
% \linenumbers

\begin{frontmatter}
\title{Self-adaptive weighting and sampling for physics-informed neural networks}

\author[1]{Wenqian Chen}
%\ead{wenqian.chen@pnnl.gov}
\author[1]{Amanda Howard}
\author[1]{Panos Stinis \corref{cor1}}
\ead{panos.stinis@pnnl.gov}

% \author[1]{...}
% \ead{...}

\cortext[cor1]{Corresponding author}

\address[1]{Advanced Computing, Mathematics and Data Division \\
	Pacific Northwest National Laboratory \\ Richland, WA 99354, USA}

%\address[1]{Advanced Computing, Mathematics and Data Division \\
%	Pacific Northwest National Laboratory \\ Richland, WA 99354, USA}

\begin{abstract}
Physics-informed deep learning has emerged as a promising framework for solving partial differential equations (PDEs). Nevertheless, training these models on complex problems remains challenging, often leading to limited accuracy and efficiency. In this work, we introduce a hybrid adaptive sampling and weighting method to enhance the performance of physics-informed neural networks (PINNs). The adaptive sampling component identifies training points in regions where the solution exhibits rapid variation, while the adaptive weighting component balances the convergence rate across training points. Numerical experiments show that applying only adaptive sampling or only adaptive weighting is insufficient to consistently achieve accurate predictions, particularly when training points are scarce. Since each method emphasizes different aspects of the solution, their effectiveness is problem dependent. By combining both strategies, the proposed framework consistently improves prediction accuracy and training efficiency, offering a more robust approach for solving PDEs with PINNs.
\end{abstract}

\begin{keyword}
Self-adaptive weighting \sep Self-adaptive sampling \sep Physics-informed neural networks
\end{keyword}

\end{frontmatter}

\section{Introduction}
Physics‐informed neural networks (PINNs) embed the governing partial differential equation (PDE) into the loss function of a deep neural network, enforcing physical consistency alongside data fidelity \cite{raissi2019physics}.  In a standard PINN formulation, the objective is composed of a data term (for initial and boundary conditions) and a PDE residual term evaluated at collocation points in the domain.  While this approach has shown promise on a variety of forward and inverse PDE problems, training can be challenging: competing loss components may have wildly different scales, and uniform sampling of collocation points often fails to focus learning on regions where the solution exhibits sharp gradients or singular behavior \cite{wang2020understanding,gao2021adaptive,wight2020solving,mcclenny2020self}.

To address the multi‐objective nature of PINN training due to competing loss terms, a range of self‐adaptive weighting strategies have been proposed. Gradient-based methods include learning rate annealing, where weights are updated inversely to back-propagated gradients \cite{wang2021understanding}. Time-structured approaches, such as causal training, assign weights in a temporally ordered fashion for time-dependent problems \cite{wang2024respecting}. Residual-based strategies are also popular: Liu and Wang \cite{liu2021dual} proposed a minimax method with gradient descent for parameters and ascent for weights; McClenny and Braga-Neto \cite{mcclenny2020self} extended this to point-wise weights; and auxiliary networks have been used for point-wise weighting \cite{song2024loss,zhang2023dasa}. Anagnostopoulos et al. \cite{anagnostopoulos2024residual} updated weights by normalized residuals. Lagrangian approaches adapt weights as multipliers for constraints, including augmented Lagrangian methods (ALM) \cite{basir2022physics}, adaptive ALM \cite{basir2023adaptive}, dual problem formulations \cite{basir2023investigating}, and point-wise multipliers \cite{son2023enhanced}. Finally, kernel-based strategies such as neural tangent kernel (NTK) weighting \cite{wang2022and,wang2022improved} and the conjugate kernel (CK) \cite{howard2024conjugate} update weights according to kernel eigenvalue information.

In parallel, adaptive sampling methods in physics-informed machine learning refine training point distributions to better capture the structure of the solution. Residual-based approaches \cite{lu2021adaptive, gao2023active} place points in regions of large residuals or select informative samples during training. Importance sampling \cite{nabian2021efficient} chooses points according to a proposal distribution derived from the loss function to improve efficiency. Residual/gradient-based strategies \cite{mao2023physics} enhance both accuracy and stability. Deep adaptive sampling \cite{tang2023pinns} employs generative models to sample high-residual regions, while annealed adaptive importance sampling \cite{zhang2025annealed} applies expectation–maximization to handle multimodal loss landscapes. For singular or sharp solutions, the expected improvement refinement \cite{liu2024adaptive} incorporates residual gradients and boundary sampling, and Gaussian mixture distribution-based sampling \cite{jiao2024gaussian} uses residual-informed distributions for adaptive selection. Collectively, these methods improve convergence and accuracy, though iterative point addition increases computational cost.

In this work, we propose a framework that integrates adaptive weighting and adaptive sampling for physics-informed machine learning. The two strategies address complementary aspects of the training process: adaptive weighting balances the contributions of different loss components, ensuring that no part of the solution dominates or is neglected, while adaptive sampling redistributes training points toward regions that are more challenging to approximate, such as areas with large residuals or sharp gradients. By combining these mechanisms, the framework provides a more comprehensive treatment of training, simultaneously stabilizing optimization and enhancing data efficiency. Numerical experiments demonstrate that the proposed method consistently achieves high prediction accuracy, highlighting the benefits of this complementary interaction.

The rest of the paper is structured as follows. Section \ref{sec_background} reviews the concept of PINNs and summarizes our previous work on adaptive weighting based on the balanced residual decay rate (BRDR) \cite{chen2025self}. Section \ref{sec_SAsampling} introduces the self-adaptive sampling method based on residuals and discusses how to combine it with the adaptive weighting method.
In Section \ref{sec_test_PINN} the proposed adaptive weighting and sampling method is tested on four benchmark problems. Finally, some conclusions will be drawn in Section \ref{sec_conclusion}.
To promote reproducibility and further research, the code and all accompanying data will be made available upon publication.

\section{Background}\label{sec_background}
\subsection{Physics-Informed Neural Networks}
Physics-informed neural networks (PINNs) are designed to approximate the solution of PDEs by minimizing a loss function that includes physics-based terms derived from the governing equations. Let’s assume we are solving a general PDE subject to boundary conditions (BCs):
\begin{align} 
	\mathcal{N}(u(\mathbf{x})) = 0, \quad \mathbf{x} \in \Omega \\
	\mathcal{B}(u(\mathbf{x})) = 0, \quad \mathbf{x} \in \partial \Omega,
\end{align}
where $\mathcal{N}$ is a differential operator and $u(\mathbf{x})$ is the solution we seek. The boundary condition is enforced by a general boundary operator $\mathcal{B}$.

The goal of the PINN is to approximate $u(\mathbf{x})$ by a neural network $u_\theta(\mathbf{x})$, where $\theta$ represents the network parameters (weights and biases). The PINN loss function usually consists of two main components,
\begin{equation}
	\mathcal{L}_{total} = \mathcal{L}_{PDE} + \mathcal{L}_{BC}.
\end{equation}
The first component is the residual loss, which ensures that the neural network approximation $u_\theta(\mathbf{x})$ satisfies the PDE:
\begin{equation}
	\mathcal{L}_{PDE} = \mathbb{E}_{\mathbf{x}\in\Omega} \left[\left|\mathcal{N}(u_\theta(\mathbf{x}))\right|^2\right] \approx
	\frac{1}{N_r} \sum_{i=1}^{N_r} \left| \mathcal{N}(u_\theta(\mathbf{x}_i)) \right|^2,
\end{equation}
where $N_r$ is the number of collocation points (randomly selected points in the domain where the PDE is enforced) and $x_i$ are the coordinates of these points. 
The second component is the boundary condition loss, which ensures that the solution satisfies the general boundary conditions imposed by $\mathcal{B}$. This loss is formulated as:
\begin{equation}
	\mathcal{L}_{BC} = \mathbb{E}_{\mathbf{x}\in\partial\Omega} \left[\left| \mathcal{B}(u_\theta(\mathbf{x}_i))  \right|^2\right] \approx
	\frac{1}{N_b} \sum_{i=1}^{N_b} \left| \mathcal{B}(u_\theta(\mathbf{x}_i))  \right|^2,
\end{equation}
where $N_b$ is the number of boundary points.

\subsection{Adaptive Weighting based on balanced residual decay rate}\label{sec_BRDR}
In our recent work \cite{chen2025self}, we introduced a self-adaptive weighting method based on the balanced residual decay rate (BRDR). This approach assigns a pointwise adaptive weight to each residual term, including both the governing equation and boundary condition residuals. The method dynamically adjusts these weights during training to balance the convergence rates across all training points, thereby improving both accuracy and efficiency in solving PDEs with PINNs. The weighted loss function is defined as follows:
\begin{align}
	\mathcal{L}(\pmb{\theta}; \mathbf{w},s) &= s \left( \dfrac{1}{N_R}\sum_{i=1}^{N_R}w_R^i\mathcal{R}^2(\mathbf{x}_R^i)
	+\dfrac{1}{N_B}\sum_{i=1}^{N_B}w_B^i\mathcal{B}^2(\mathbf{x}_B^i)
	\right)	\label{eq_loss_weighted}\\
	&s.t. \qquad  \text{mean}(\mathbf{w}):=\frac{\sum_{i=1}^{N_R}w_R^i+\sum_{i=1}^{N_B}w_B^i}{N_R+N_B}=1 \label{eq_weight_constraint}
\end{align}
where $w_R^i>0$ is the weight assigned to the residual of each residual collocation point, $w_B^i>0$ is the weight assigned to each boundary point, and $\mathbf{w}$ is the collection of these weights. 
The scale factor $s$ is employed to scale all the weights, so that the formulation could cover all kinds of possible weight distributions. 

The basic idea behind BRDR is to update weights based on two critical observations:
\begin{enumerate}
	\item Residuals at different training points may vary significantly across the domain.
	\item The point with the smallest residual decay rate often dominates the convergence speed of the global solution.
\end{enumerate}

To quantify the speed of residual decay, we use the inverse residual decay rate (IRDR), defined as:
\begin{equation}
	\text{IRDR} = \frac{R^2(t)}{\sqrt{\overline{R^4}(t) + \epsilon}}
\end{equation}
where $ R(t) $ represents the residual at iteration $ t $, $ \overline{R^4}(t) $ is the exponential moving average of $ R^4(t) $, and $ \epsilon $ is a small constant to avoid division by 0. The exponential moving average $ \overline{R^4}(t) $ is updated using:

\begin{equation}
	\overline{R^4}(t) = \beta_c \, \overline{R^4}(t-1) + (1 - \beta_c) \, R^4(t)
\end{equation}
where $ \beta_c $ is a smoothing constant that controls the influence of past residuals.

Since a larger IRDR indicates a slower residual decay, we assign higher weights to loss terms with larger IRDR values. To manage these weights dynamically during training, we compute reference weights at each iteration $ t $ based on the relative IRDR values with respect to their global mean. This strategy ensures that the mean of the weights remains at 1, keeping the weights bounded throughout the training process:

\begin{equation}
	\mathbf{w}_t^{\text{ref}} = \frac{\mathbf{IRDR}_t}{\text{mean}(\mathbf{IRDR}_t)}
\end{equation}
where $ \mathbf{IRDR}_t $ is the vector of IRDR values for all the training items at iteration $ t $.

To minimize noise and stabilize weight updates during training, the weights are adjusted using an exponential moving average:

\begin{equation}
	\mathbf{w}_t = \beta_w \, \mathbf{w}_{t-1} + (1 - \beta_w) \, \mathbf{w}_t^{\text{ref}}
\end{equation}
where $ \beta_w $ is a smoothing factor that helps to smooth out fluctuations in the weights.
For more details on the BRDR method, please refer to our previous work \cite{chen2025self}.

\section{Adaptive Training in Physics-Informed Machine Learning}
While adaptive weighting adjusts the importance of different training points, it often requires a high density of points in regions with large gradients, leading to a large number of residual points when using random sampling and increasing significantly the computing cost for training. To reduce the number of training points, we develop an adaptive sampling method so that we can use fewer training points while achieving a higher training accuracy. 

\subsection{Issues with overfitting}\label{sec_overfitting}
Here we take a 1D perturbation equation as an example to showcase the possible issues that can arise when using the adaptive weighting method with a small number of training points. The 1D perturbation equation we use as an example is given by
\begin{equation}
	\begin{aligned}
		-\epsilon^2 \frac{d^2u}{dx^2}+u(x)&=1, \qquad x\in [0,1] \\
		u(0)=u(1)&=0
	\end{aligned}
\end{equation}	
where the parameter $\epsilon$ is set to $10^{-4}$. The analytical solution of this problem is given by:
\begin{equation}
	u(x) = 1-\frac{e^{-x/\epsilon}+e^{(x-1)/\epsilon}}{1+e^{-1/\epsilon}}.
\end{equation}
The solution contains two thin boundary layers at $x=0$ and $x=1$, with the boundary layer thickness is  proportional to $\epsilon$. The stiff gradient within the boundary layers presents a challenge to PINN training.

\begin{figure}[!ht]
	\centering	
	\includegraphics[ width=14cm, trim=3cm 0cm 3.0cm 0cm, clip=true]{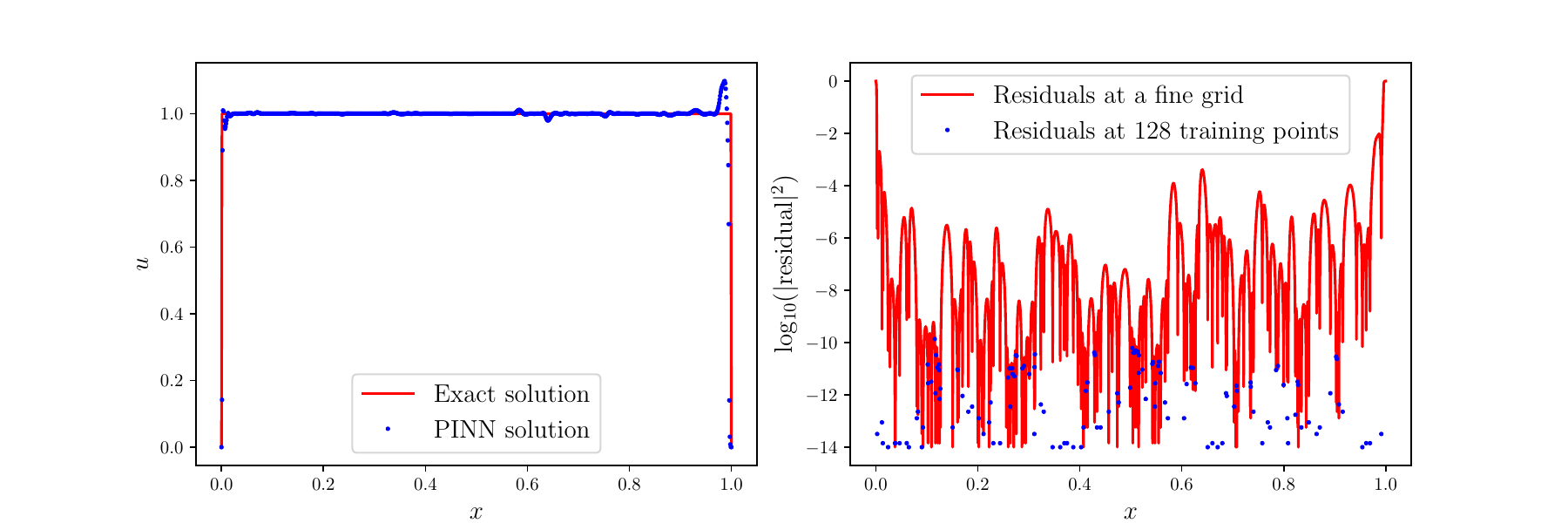}
	\caption{(Left) PINN prediction for the 1D perturbation equation, and (right) residuals at the training points compared with those on a fine grid.
	}
	\label{Fig_Weight_Sample_residuals}	
\end{figure}

To solve this problem with PINNs, we uniformly sample 128 training points within the domain. The PINN prediction is shown in Fig. \ref{Fig_Weight_Sample_residuals}. The residuals at the training points are minimized to a very small magnitude (less than $10^{-10}$), yet the PINN prediction shows a large deviation from the ground truth. This discrepancy arises because the residuals at unseen points remain significant. In other words, the fixed training points underestimate the average residuals, which ultimately leads to the large error of the PINN prediction.

\subsection{Self-adaptive sampling based on residuals}\label{sec_SAsampling}

\begin{figure}[htbp]
	\centering	
	\includegraphics[ width=14cm, trim=0cm 0cm 0cm 0cm, clip=true]{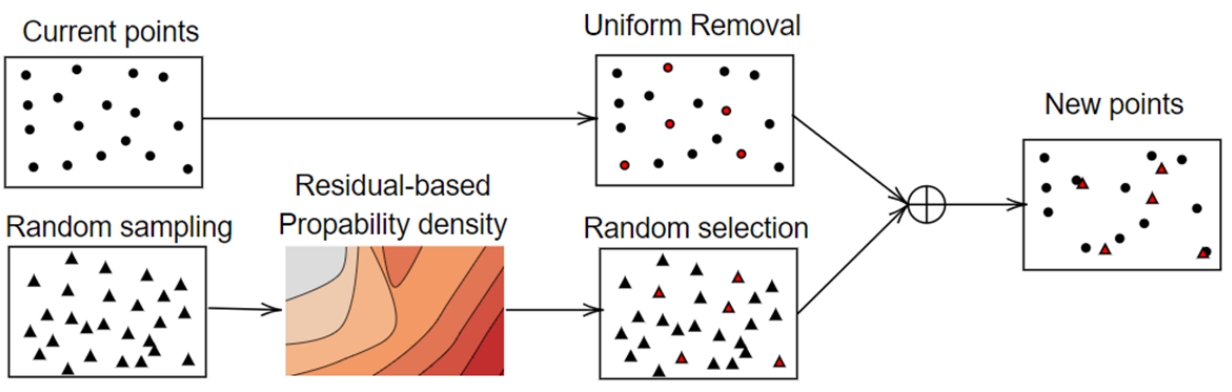}
	\caption{Schematic illustration of the implementation process for self-adaptive sampling based on residuals.
	}
	\label{Fig_sampling_method}	
\end{figure}

Considering the overfitting issue discussed in Section \ref{sec_overfitting}, a straightforward remedy is to introduce additional training points in regions where the residuals are large. To achieve this, we first compute the residuals $R(x)$ over a set of candidate points. New training points are then randomly selected from these candidates according to a probability distribution, as shown in Fig. \ref{Fig_sampling_method}. The probability of selection is proportional to the square of the corresponding residuals:

\begin{equation}
	p(x) = \frac{R^2(x)}{\sum_{x} R^2(x)}
\end{equation}

To prevent overemphasizing points with extremely large residuals and underemphasizing those with extremely small residuals, we clip the residuals using:

\begin{equation}
	R_{\text{clipped}}^2(x) = \max \left( \hat{R}^2, \min \left( R^2(x), \gamma \hat{R}^2 \right) \right)
\end{equation}
where $ \hat{R}^2 $ is the median value of $ R^2(x) $ over the candidate points, and $ \gamma = 100 $ is a scaling factor unless stated otherwise.

Resampling is performed after a specified number of optimization steps. To prevent oscillations caused by abruptly replacing all training points, only a fraction $p_u$ of the existing points are updated in each resampling cycle, while the remaining points are retained.

% Additionally, as indicated in the literature \cite{chen2025self}, the prediction error decreases rapidly during the initial training stage, where the model captures the large-scale solution structure first. Therefore, it is beneficial to keep the residual points unchanged during the initial phase of training. 

\subsection{Combination of Adaptive Weighting and Sampling}

To effectively combine adaptive weighting with adaptive sampling, it is necessary to determine the weights, or weight-related quantities, for the newly introduced training points. As described in Section \ref{sec_BRDR}, both the weights and $\overline{R^4}$ are updated at every training step. When new points are added through adaptive sampling, their corresponding weights and $\overline{R^4}$ must also be computed.

To estimate these quantities at the new points, we employ inverse distance weighting (IDW) interpolation. This interpolation allows us to estimate the weights and $ \overline{R^4} $ for new points based on the existing training points, ensuring a smooth transition in the weighting scheme when training points are updated.

\begin{equation}
	w_{\text{new}} = \frac{\sum_{i} \dfrac{w_i}{r_i}}{\sum_{i} \dfrac{1}{r_i}}
	,\qquad
	\overline{R^4}_{\text{new}} = \frac{\sum_{i} \dfrac{\overline{R^4}_i}{r_i}}{\sum_{i} \dfrac{1}{r_i}}
\end{equation}

where:
\begin{itemize}
	\item $ w_{\text{new}} $ and $ \overline{R^4}_{\text{new}} $ are the weight and exponential moving average of $ R^4 $ at the new point,
	\item $ w_i $ and $ \overline{R^4}_i $ are the corresponding quantities at old points,
	\item $ r_i $ is the distance between the new point and the $ i $-th old point.
\end{itemize}

The combined algorithm is summarized in Algorithm \ref{alg_combination}.
For adaptive weighting, key parameters include $\beta_c$ (set to $0.999$) for smoothing the exponential moving average of $R^4$ and $\beta_w$ (set to $0.999$) for weight updates, both of which perform well across various PINN problems \cite{chen2025self}.
For adaptive sampling, $\gamma$ (set to $100$) controls residual clipping, $p_u$ (set to $0.2$) is the fraction of points updated per resampling, and $N_s$ (set to $100$) determines the update frequency, namely the interval between resampling steps.
Default values are found empirically to be  effective but may be tuned for specific problems; detailed analyses are provided in \ref{sec_hyperparameter}.

\begin{algorithm}[H] 
	\caption{Adaptive Weighting and Sampling}
	\begin{algorithmic}[1]
		\Require Initial training points $ \mathcal{X}_0 $
		\For{each training iteration $ t = 1, 2, \dots, T $}
		\State \textbf{Adaptive Weighting:}
		\State Compute residuals $ R_t(x) $ at current training points $ \mathcal{X}_{t-1} $
		\State Update $ \overline{R^4}_t(x) = \beta_c \, \overline{R^4}_{t-1}(x) + (1 - \beta_c) \, R_t^4(x) $
		\State Calculate IRDR: $ \text{IRDR}_t(x) = \dfrac{R_t^2(x)}{\sqrt{\overline{R^4}_t(x) + \epsilon}} $
		\State Compute reference weights: $ w_t^{\text{ref}}(x) = \dfrac{\mathbf{IRDR}_t(x)}{\text{mean}(\mathbf{IRDR}_t(x))} $
		\State Update weights: $ w_t(x) = \beta_w \, w_{t-1}(x) + (1 - \beta_w) \, w_t^{\text{ref}}(x) $
		\State \textbf{Adaptive Sampling:}
		\If{$ t \mod N_s = 0 $}		
		\State Generate candidate points $ \mathcal{X}_{\text{cand}} $
		\State Compute residuals $ R_t(x) $ for $ x \in \mathcal{X}_{\text{cand}} $
		\State Compute selection probabilities with clipped residuals:
		\begin{equation*}
			p(x) = \dfrac{R_{\text{clipped}}^2(x)}{\sum_{x \in \mathcal{X}_{\text{cand}}} R_{\text{clipped}}^2(x)}, \qquad \text{where} \quad 			R_{\text{clipped}}^2(x) = \max \left( \hat{R}^2, \min \left( R_t^2(x), \gamma \hat{R}^2 \right) \right)
		\end{equation*}
		\State Select $p_u|\mathcal{X}_{t-1}|$ new points $ \mathcal{X}_{\text{new}} $ based on $ p(x) $
		\State Select $p_u|\mathcal{X}_{t-1}|$ old points $ \mathcal{X}_{\text{replace}} $ randomly.		
		\State \textbf{IDW Interpolation for New Points:}
        \State $w_t(\mathcal{X}_{\text{new}})$ = {IDW}($w_t(\mathcal{X}_{t-1})$, $\mathcal{X}_{t-1}$, $\mathcal{X}_{\text{new}}$ )
        \State $\overline{R^4}_t(\mathcal{X}_{\text{new}})$ = {IDW}($\overline{R^4}_t(\mathcal{X}_{t-1})$, $\mathcal{X}_{t-1}$, $\mathcal{X}_{\text{new}}$ )
		\State Update training points: $ \mathcal{X}_{t} = (\mathcal{X}_{t-1} \setminus \mathcal{X}_{\text{replace}}) \cup \mathcal{X}_{\text{new}} $        
		\EndIf
		\State \textbf{Assemble the total loss and perform backward propagation}
		\State \textbf{Update the parameters with gradient descent}
		\EndFor
	\end{algorithmic}
	\label{alg_combination}
\end{algorithm}

\section{Numerical results}\label{sec_test_PINN}
To validate the performance of the proposed self-adaptive weighting and sampling approach in training PINNs, we test on four benchmark problems: the perturbation equation, the Allen–Cahn equation, the Burgers equation, and the lid-driven cavity flow. The prediction accuracy is  assessed using the $L_2$ relative error defined as

\begin{equation}
\epsilon_{L_2} = \frac{\|u - u_E\|_2}{\|u_E\|_2}
\end{equation}
where $u$ and $u_E$ are the vectors of the predicted and reference solutions on the test set, respectively.

In our experiments, we employ the mFCN network architecture (see \ref{sec_net}) which consists of six hidden layers with 128 neurons each. The hyperbolic tangent function is utilized as the activation function throughout the network. Network parameters are initialized using Kaiming Uniform initialization \cite{he2015delving}; specifically, for a layer with shape (out\_features, in\_features), the weights and biases are sampled from $\mathcal{U}\Big(-\sqrt{k}, \sqrt{k}\Big)$ with $k = 1/\text{in\_features}$. All implementations are carried out in PyTorch \cite{paszke2019pytorch} and executed on a GPU cluster using a 32-bit single-precision data type on an NVIDIA\textsuperscript{\textregistered} Tesla P100 GPU. 

\begin{figure}[htbp]
	\centering	
	\includegraphics[width=16cm, trim=1.0cm 0.1cm 2.0cm 0cm, clip=true]{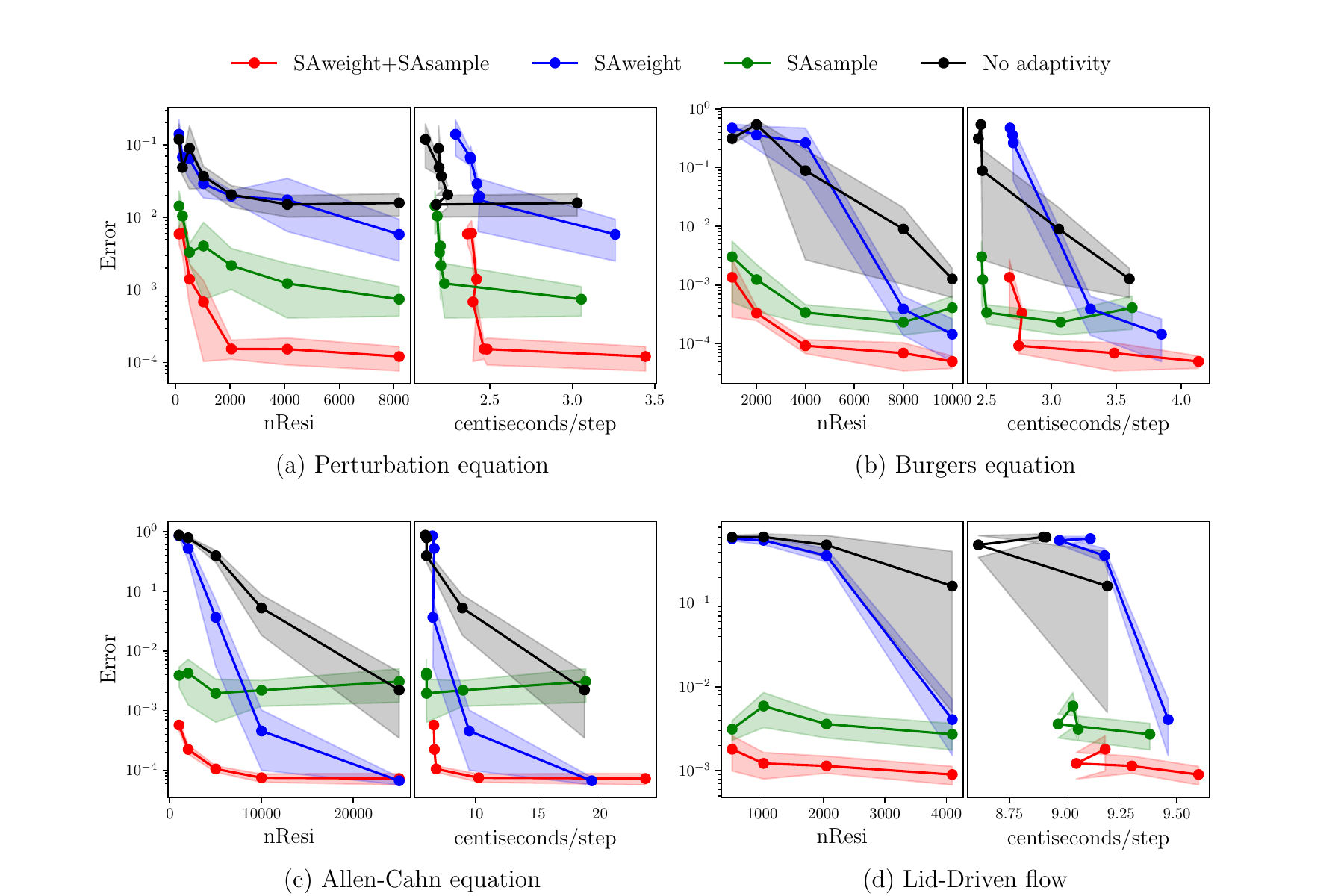}
	\caption{PINN prediction errors (left panels) and training time cost (right panels) for the perturbation, Allen–Cahn, Burgers’ equations, and lid-driven flow. Shaded areas denote the mean ± 2 standard deviations, calculated from 5 independent runs for each case.
	}
	\label{Fig_ErrorCost_PINN}	
\end{figure}

To assess the impact of adaptive sampling and weighting strategies, Fig. \ref{Fig_ErrorCost_PINN} presents the PINN prediction errors and training costs for the four representative problems. Across all training methods, increasing the number of residual points generally reduces the prediction error--especially for adaptive weighting--though the rate of improvement varies notably among methods and problems. For the perturbation and Burgers equations, the error decreases steadily as the number of residual points (i.e., the batch size) increases. In contrast, for the Allen–Cahn equation and  the lid-driven cavity flow, the error under adaptive sampling remains nearly unchanged with larger batch sizes. These results demonstrate that the effectiveness of adaptive sampling and weighting is strongly problem dependent, making it difficult to rely solely on either strategy for robust performance. By contrast, the combined approach consistently achieves the best accuracy across all tested cases.

In terms of computational cost, the combined approach increases training time by less than 20\% relative to the non-adaptive PINN under the same batch size and maximum iteration settings. More importantly, for a given target prediction error, the proposed method achieves results with substantially less training time than the other strategies.

To visualize the distribution of training points and their associated weights for the Allen–Cahn, Burgers, and lid-driven cavity flow problems, we plot scatter diagrams of the training points colored by their weights, as shown in Figs. \ref{Fig_distribution_AllenCahn}–\ref{Fig_distribution_LidDriven} in  \ref{sec_distribution}. 
While these plots allow for some preliminary observations, they do not clearly reveal the focus of different methods, particularly for the combined approach. To provide a more quantitative assessment, we employ kernel density estimation (KDE) to evaluate the relative importance of different regions. Specifically, we use \textit{gaussian\_kde} in SciPy, which by default applies Scott’s rule for bandwidth selection \cite{scott1992multivariate}, and incorporate the adaptive weights by assigning them as kernel weights. The detailed results will be discussed in the following subsections.

\begin{table}[htbp]
	\centering
	\caption{The choice of location of the training points and the BRDR training setup for solving different problems with PINNs.}	
	\begin{tabular}{ccccc}
		\toprule
		Problems           & Allen–Cahn                                                                 & Perturbation                                                                   & Burgers & Lid-Driven\\ 
		\midrule
		PDE points         & Latin Hypercube          & Unifrom            &  Latin Hypercube    & Random \\ \hline
		IC points          & Uniform            & --                                                                          & Uniform      &  --      \\ \hline
		BC points          & --                                                                         &Uniform                 &Random  &Uniform  \\ \hline
		Network            & \begin{tabular}[c]{@{}c@{}}{[}21{]}+{[}128{]}$\times$6+{[}1{]}\\ mFCN \\tanh\end{tabular} & \begin{tabular}[c]{@{}c@{}}{[}2{]}+{[}128{]}$\times$6+{[}1{]}\\ mFCN \\ tanh\end{tabular} & \begin{tabular}[c]{@{}c@{}}{[}2{]}+{[}128{]}$\times$6+{[}1{]}\\ mFCN \\tanh\end{tabular} &\begin{tabular}[c]{@{}c@{}}{[}2{]}+{[}128{]}$\times$6+{[}3{]}\\ mFCN \\tanh\end{tabular} \\ \hline
		Adam steps   & 3e5                                                                    & 1e5                  &  4e4  &8E4 \\ \hline
		Adam Learning rate      & $0.001\times0.99^{t/750}$                                      & $0.005\times0.99^{t/250}$                                       &    $0.001\times0.99^{t/100}$  &    $0.001\times0.99^{t/400}$     \\ \hline
		($\beta_c, \beta_w$,) in BRDR & (0.999, 0.999)                                                             & (0.999, 0.999)                                                             &  (0.999, 0.999)    &(0.999, 0.999)     \\
		($p, \gamma, N_s$) in SAAR & (0.2, 100, 100)                                                             & (0.2, 100, 100)                                                             &  (0.2, 100, 100)   &(0.2, 100, 100)       \\		
		\bottomrule     
	\end{tabular}
	\label{tab_setup_PINN}
\end{table}

\subsection{Perturbation problem}
We revisit the 1D perturbation problem introduced in Section \ref{sec_overfitting} to evaluate the effectiveness of the proposed adaptive weighting and sampling strategies. The training setup is summarized in Table \ref{tab_setup_PINN}. Figure \ref{Fig_ErrorCost_PINN} presents the prediction accuracy of PINN training with and without adaptive weighting and sampling. For this problem, adaptive weighting alone provides little improvement, whereas adaptive sampling significantly enhances training accuracy. The combination of adaptive weighting and sampling yields the best performance. In particular, the combined approach achieves an $L_2$ relative prediction error of 0.01\% with 2048 residual points. For comparison, the prediction error reported in Ref. \cite{fang2023ensemble} is 0.43\%, and the standard PINN produces an error of about 12.56\%, both using a much larger batch size of 10,000.

\begin{figure}[htbp]
	\centering	
	\includegraphics[ width=10cm, trim=1.0cm 0.2cm 1.0cm 1.2cm, clip=true]{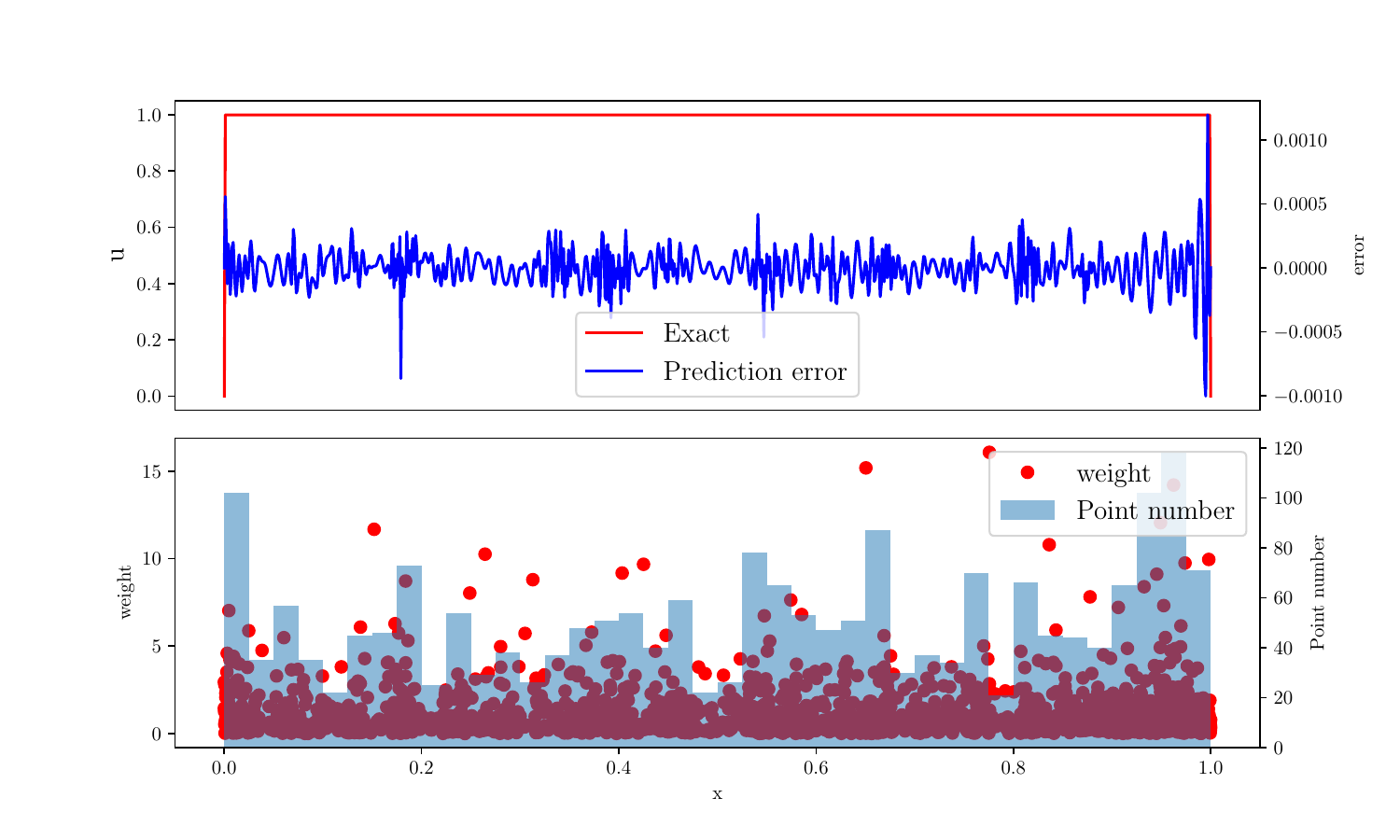}
	\caption{Perturbation problem: (top) PINN prediction errors and exact solution and (bottom) weight distribution and residual point distribution.
	}
	\label{Fig_distribution_perturbation}	
\end{figure}

Figure \ref{Fig_distribution_perturbation} shows the pointwise PINN prediction error together with the weight and residual-point distributions obtained from the combined adaptive weighting and sampling strategy with the batch size 2048. The PINN prediction closely matches the exact solution, even within the thin boundary layers, where the absolute pointwise error remains below $10^{-3}$. Residual points are more densely concentrated in these regions, demonstrating that adaptive sampling effectively allocates training efforts where they are most needed. Meanwhile, the weight distribution remains relatively uniform across the domain, indicating that adaptive weighting successfully balances the contribution of each training point according to its residual decay rate. This homogeneous pattern confirms that the method mitigates the dominance of slow-converging points and promotes consistent convergence, consistent with the findings of the previous study \cite{chen2025self}.

\subsection{Allen-Cahn equation}
The Allen-Cahn equation is defined as follows:
\begin{equation}
	\begin{aligned}
		\frac{\partial u}{\partial t} -5(u-u^3)-D \frac{\partial^2 u}{\partial x^2} &= 0, \qquad (x,t)\in [-1,1]\times[0,1] \\
		u(x,0)&=x^2 \cos(\pi x), \qquad x\in[-1,1] \\
		u(\pm 1,t)&=0,        \qquad \qquad t\in[0,1]  \\
	\end{aligned}
\end{equation}
where the viscosity $D=1E-4$ is considered. 

As adopted in reference \cite{anagnostopoulos2024residual,chen2025self}, we use a Fourier feature transformation on $x$ to enhance the network model's ability to approximate periodic functions. While the Allen-Cahn equation does not explicitly have periodic boundary conditions, this transformation helps improve the model's expressiveness. With 10 Fourier modes, the two-dimensional input $\mathbf{x} = (x, t)$ is expanded into a 21-dimensional feature vector $\widehat{\mathbf{x}}$, which is then fed into the network through the following mapping:
\begin{equation}
	\label{eq_featureTransform}
	\widehat{\mathbf{x}} = \gamma(\mathbf{x}) =
	\left[
	\sin(\pi \mathbf{B} x),
	\cos(\pi \mathbf{B} x),
	t
	\right]^T, 
\end{equation}
where $\mathbf{B} = [1, \ldots, 10]^T$.

Figure \ref{Fig_ErrorCost_PINN} illustrates the prediction error from PINN training with and without adaptive weighting and sampling. For this problem, increasing the batch size under adaptive sampling alone yields no significant improvement. In contrast, adaptive weighting leads to a clear reduction in training error as the batch size increases. The combined use of adaptive weighting and sampling consistently achieves the lowest prediction error.

% Figure \ref{Fig_distribution_AllenCahn} shows the weight and residual-point distributions from the combined adaptive weighting and sampling strategy. Since the prediction error is as small as $2.28E-5$, the prediction itself is not shown. The distribution of residual points is nearly uniform and does not exhibit a clear correlation with the exact solution, indicating that adaptive sampling provides little benefit in this case and the network could handle the large-gradient area effectively. Instead, adaptive weighting plays a more critical role in improving prediction accuracy.

\begin{figure}[htbp]
	\centering	
	\includegraphics[ width=15cm, trim=1.0cm 0.2cm 2.0cm 1.2cm, clip=true]{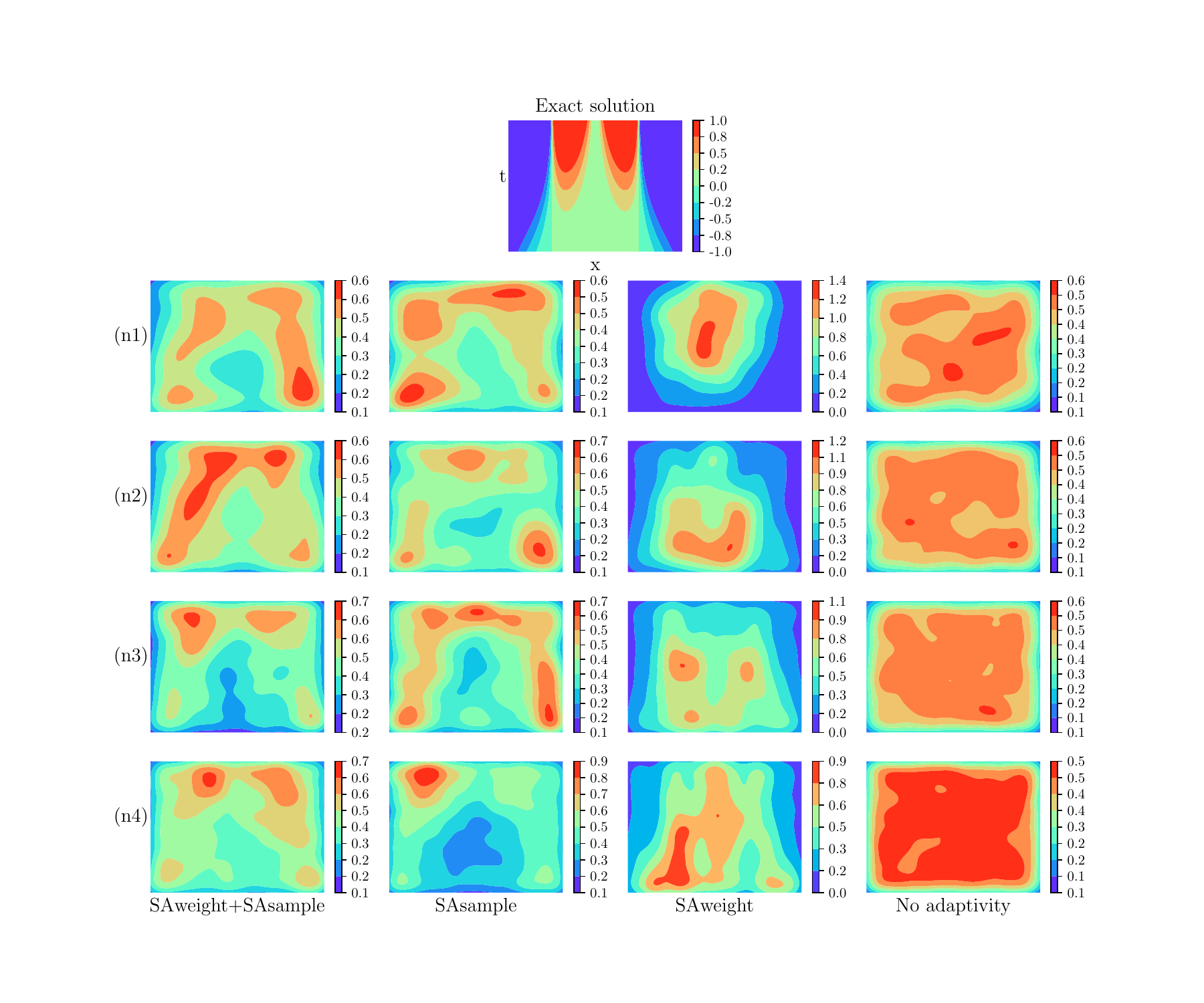}
	\caption{Weighted density estimation for Allen-Cahn equation. ``n1-n4" denotes the number of training points 2000, 5000, 10000, 25000, respectively.
	}
	\label{Fig_distributionKDE_AllenCahn}	
\end{figure}

Figure \ref{Fig_distributionKDE_AllenCahn} shows the weighted density estimation for the Allen–Cahn equation. With adaptive sampling, the training points become more concentrated in large-gradient regions, enabling more effective learning, particularly for smaller batch sizes. In contrast, solely adaptive weighting fails to identify these large-gradient regions when the batch size is small, requiring larger batch sizes to be effective. Moreover, adaptive weighting places greater emphasis on the initial condition, as indicated by the roughly decreasing density along the $t$ dimension for batch sizes $n2$–$n4$. While adaptive sampling effectively captures large-gradient regions, the combined strategy blends the strengths of both methods, focusing simultaneously on the initial condition and the large-gradient regions, thereby achieving improved prediction accuracy.

\subsection{Burgers equation}
\label{sec_Burgers}
The Burgers equation is defined as follows:
\begin{equation}
	\begin{aligned}
	\frac{\partial u}{\partial t} + u\frac{\partial u}{\partial x} -v \frac{\partial^2 u}{\partial x^2} &= 0, \qquad (x,t)\in [-1,1]\times[0,1] \\
	u(x,0)&=-\sin(\pi x), \qquad x\in[-1,1] \\
	u(\pm 1,t)&=0,        \qquad \qquad t\in[0,1]  \\
	\end{aligned}
\end{equation}
where $u$ is the flow velocity, and the viscosity of the fluid $v=0.01/\pi$ is considered.

Figure \ref{Fig_ErrorCost_PINN} illustrates the  prediction accuracy from PINN training with/without adaptive weighting and with/without adaptive sampling. For this problem,  no significant improvements from increasing batch size is gained from solely adaptive sampling. On the contrary, the training error of adaptive weighing decreases with the increase of batch size. The combination of adaptive weighting and adaptive sampling consistently achieves the lowest prediction error.

\begin{figure}[htbp]
	\centering	
	\includegraphics[ width=15cm, trim=1.0cm 0.2cm 2.0cm 1.2cm, clip=true]{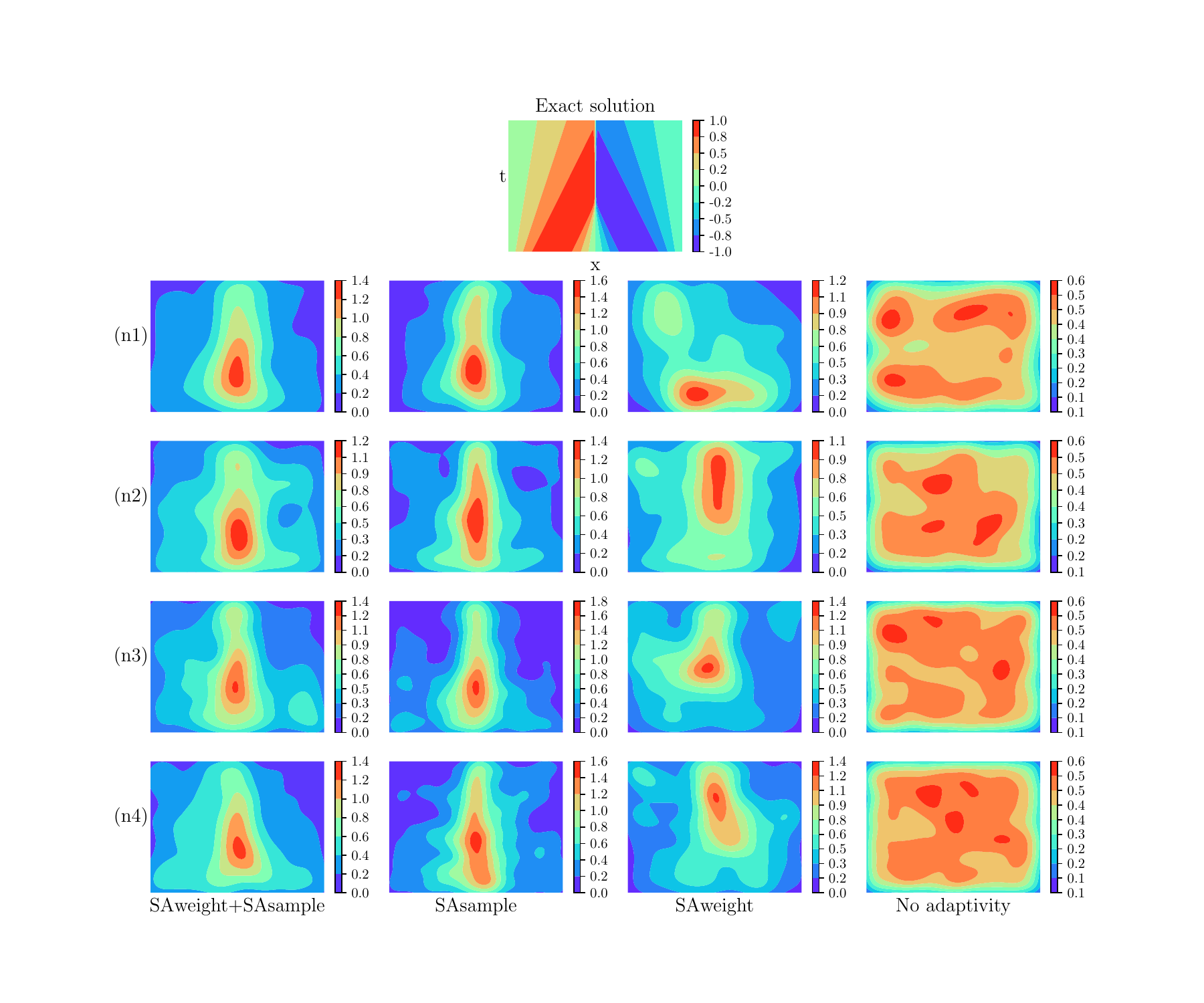}
	\caption{Weighted density estimation for Burgers equation. ``n1-n4" denotes the number of training points 2000, 4000, 8000, and 10000, respectively. 
	}
	\label{Fig_distributionKDE_Burgers}	
\end{figure}

Figure \ref{Fig_distributionKDE_Burgers} presents the weighted density estimation for the Burgers’ equation. For solely adaptive sampling, the density pattern—concentrated near the shock region—remains largely unaffected by batch size. In contrast, under solely adaptive weighting, the density distribution varies considerably with batch size: when the batch size is 2000, the density is concentrated near the initial condition, but as the batch size increases, it gradually shifts toward the shock region. Compared with solely adaptive sampling, the combined approach places greater emphasis on the initial condition, an effect inherited from adaptive weighting, which ultimately leads to improved prediction accuracy.

\subsection{Steady lid-driven cavity flow problem}
{\label{subsec_lid_steady}}
In this subsection, the lid-driven cavity flow problem is used to test the performance of the proposed MF approach and also the influence of several hyperparameters. The  flow  enclosed in a square cavity $\Omega=[0,1]^2$ is described by the (non-dimensional) incompressible Navier-Stokes equations
\begin{equation}
\left \{
\begin{aligned}
\nabla \cdot {\mathbf{u}} &= 0,  &\mathbf{x} &\in \Omega\\
 {\mathbf{u}} \cdot \nabla {\mathbf{u}} &=  - \nabla p + \frac{1}{Re}{\nabla^2}{\mathbf{u}}, & \mathbf{x} &\in \Omega \\
\mathbf{u}(\mathbf{x})&=(u_w(\mathbf{x}), 0),  & \mathbf{x} &\in \Gamma _1  \\
 \mathbf{u}(\mathbf{x})&=0,     & \mathbf{x} & \in \Gamma _2 \\
\end{aligned}
\right .
,
\label{eq_GoverningEqsLidDriven}
\end{equation}
where $\mathbf{u}=(u,v)$ is the velocity in the Cartesian coordinate system $\mathbf{x}=(x,y)$, $p$ is the pressure and $Re$ is the Reynolds number. The boundary is $\partial \Omega=\Gamma _1 \cup \Gamma _2$, where $\Gamma _1$ represents the top moving lid and $\Gamma _2$ represents the other three static non-slip walls.
$u_w$ is the driving velocity of the moving lid. To overcome the  singularity at the two upper corner points where the moving lid meets the two stationary vertical walls, a zero-corner-velocity profile $u_w$ is employed \cite{chen2018multidomain}:
\begin{equation}
u_w(\mathbf{x}) = 16x^2(1-x)^2
\end{equation}

Figure \ref{Fig_ErrorCost_PINN} illustrates the  prediction accuracy from PINN training with/without adaptive weighting and with/without adaptive sampling. For this problem,  no significant improvements from increasing batch size is gained from solely adaptive sampling. On the contrary, the training error of adaptive weighting decreases with the increase of batch size. The combination of adaptive weighting and adaptive sampling consistently achieves the lowest prediction error.

\begin{figure}[htbp]
	\centering	
	\includegraphics[ width=15cm, trim=1.0cm 0.2cm 2.0cm 1.2cm, clip=true]{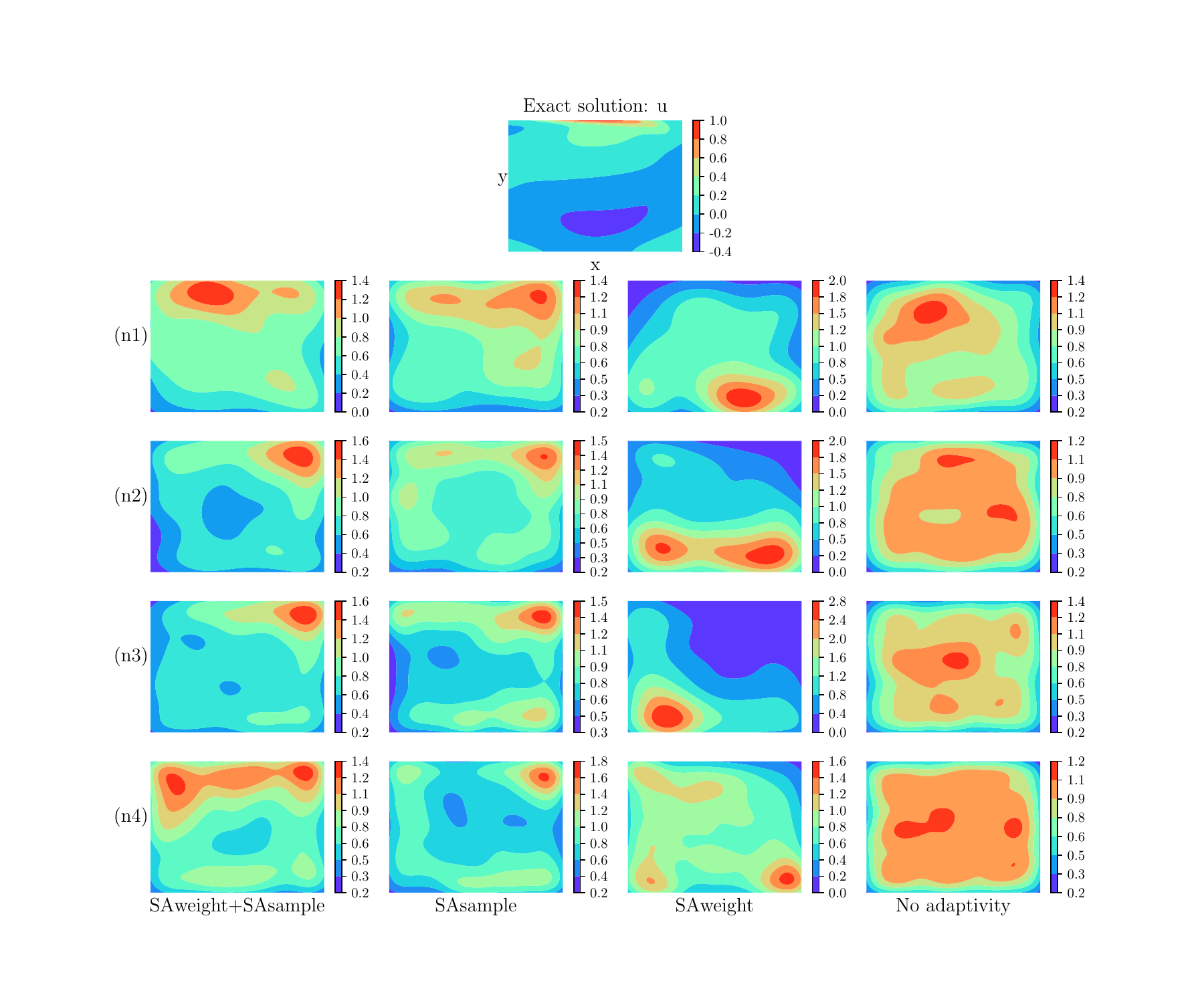}
	\caption{Weighted density estimation for the lid-driven flow. ``n1-n4" denotes the number of training points 512, 1024, 2048, 4096, respectively.
	}
	\label{Fig_distributionKDE_LidDriven}	
\end{figure}

Figure \ref{Fig_distributionKDE_LidDriven} shows the weighted density estimation for the lid-driven cavity flow problem. Across different batch sizes, the solely adaptive sampling method concentrates on the top-right corner, where the moving lid meets the stationary wall and generates large gradients. In contrast, adaptive weighting places greater emphasis on the two bottom corners, where resolving the secondary vortices is crucial for accurately capturing the flow structures. The combined strategy effectively incorporates both aspects, focusing on all critical regions and thereby improving overall performance.

\section{Conclusion}\label{sec_conclusion}
In plain PINNs, the loss function is defined as a linear combination of the squared residuals from the PDEs and boundary conditions evaluated at a set of training points. However, plain PINNs often struggle with problems involving large gradients. Adaptive weighting methods have been shown to mitigate this issue, but their effectiveness is limited, particularly when the training points are insufficient to resolve large-gradient regions. Adaptive sampling offers a natural complement by directing training points toward regions of the domain where the structure of the solution is more difficult to capture accurately.

In this work, we propose a self-adaptive sampling method based on residuals as a complement to our previously developed adaptive weighting approach. In the combined framework, more training points are allocated to regions with large gradients, while larger weights are assigned to points where residuals decay slowly. The effectiveness of the proposed strategy is evaluated on four benchmark problems. The results show that neither adaptive weighting nor adaptive sampling alone is sufficient to ensure robust performance across all problems. By contrast, their combination consistently yields superior prediction accuracy.

\section*{Acknowledgments}
The work is  supported by the U.S. Department of Energy, Advanced Scientific Computing Research
program, under the Scalable, Efficient and Accelerated Causal Reasoning Operators, Graphs and Spikes for Earth and
Embedded Systems (SEA-CROGS) project (Project No. 80278).
Pacific Northwest National Laboratory (PNNL) is a multi-program national laboratory operated for the U.S. Department of Energy (DOE) by Battelle Memorial Institute under Contract No. DE-AC05-76RL01830.

\appendix
\section{Network architectures}\label{sec_net}
The modified fully-connected network (mFCN) is introduced in the reference\cite{wang2021understanding}, and has demonstrated to be more effective than the standard fully-connected neural network.
A mFCN maps the input $\mathbf{x}$ to the output $\mathbf{y}$. Generally, a mFCN consists of an input layer, $L$ hidden layers and an output layer.  The $l$-th layer has $n_l$ neurons, where $l=0,1,..L,L+1$ denotes the input layer, first hidden layer,..., $L$-th hidden layer and the output layer, respectively. Note that the number of neurons of each hidden layer is the same, i.e., $n_1=n_2=...=n_L$. The forward propagation, i.e. the function $\mathbf{y}=f_{\pmb{\theta}}(\mathbf {x})$,  is defined as follows
\begin{equation}
	\begin{aligned}
		\mathbf{U}&= \phi(\mathbf{W}^U\mathbf{x}+\mathbf{b}^U) &&\\
		\mathbf{V}&= \phi(\mathbf{W}^V\mathbf{x}+\mathbf{b}^V) &&\\
		\mathbf{H}^{1} &= \phi( \mathbf{W}^{1}\mathbf{x}+ \mathbf{b}^{1}) &&\\
		\mathbf{Z}^{l} &= \phi( \mathbf{W}^{l}\mathbf{H}^{l-1}+ \mathbf{b}^{l}), && 2\le l \le L \\
		\mathbf{H}^{l} &= (1-\mathbf{Z}^{l}) \odot \mathbf{U} + \mathbf{Z}^{l} \odot \mathbf{V}, && 2\le l \le L \\
		f_{\pmb{\theta}}(\mathbf{x})   &=\mathbf{W}^{L+1}\mathbf{H}^{L}+ \mathbf{b}^{L+1} &&\\
	\end{aligned}
	,
\end{equation}
where  $\phi(\bullet)$ is a point-wise activation and $\odot$ denotes point-wise multiplication. The training parameter in the network is $\pmb{\theta}=\{\mathbf{W}^U,\mathbf{W}^V, \mathbf{b}^U, \mathbf{b}^V, \mathbf{W}^{1:L+1}, \mathbf{b}^{1:L+1} \}$.

\section{Hyperparameter study for adaptive sampling}\label{sec_hyperparameter}
The adaptive sampling method involves three key hyperparameters: the fraction of points to be updated $p_u$, the clipping parameter $\gamma$, and the update frequency $N_s$. To investigate their influence on training performance, we conduct a hyperparameter study using the Burgers equation as a test case, as described in Section \ref{sec_Burgers}. To study the effect of each hyperparameter independently, we vary one while keeping the others fixed at $p_u=0.2$, $\gamma=100$, and $N_s=100$. The other training settings are consistent with those in Section \ref{sec_Burgers}. Note that all the training cases share the same random seed to ensure that the initial network parameters and the initial set of residual points are identical.

\begin{figure}[htbp]
	\centering	
	\subfigure{
		\includegraphics[ width=15cm]{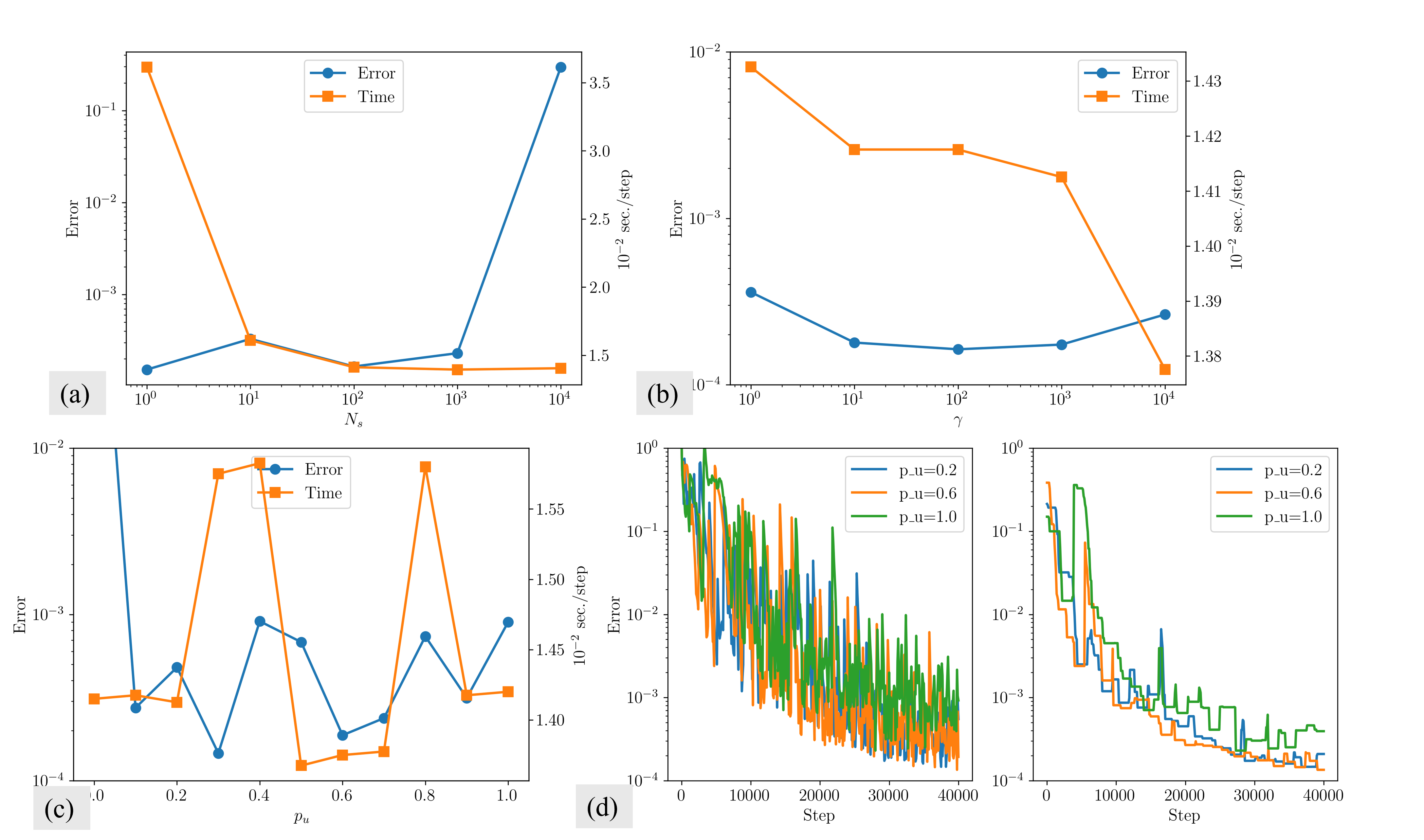}
	}	
	\caption{Hyperparameter study of adaptive sampling for the Burgers equation. The prediction error and training time per step are shown as functions of (a) the update frequency $N_s$, (b) the clipping parameter $\gamma$, and (c) the fraction of points updated $p_u$. Subfigure (d) compares the raw (left) and smoothed (right) prediction error histories for different $p_u$. The smoothing is performed using a moving minimum filter with a window size of 1500 training steps to improve visual clarity and facilitate comparison.
	}
	\label{Fig_Hyperparameter_Burgers}
\end{figure}

The prediction errors as well as the training time cost for different hyperparameter values are presented in Fig. \ref{Fig_Hyperparameter_Burgers}. For the update frequency $N_s$ shown in Fig. \ref{Fig_Hyperparameter_Burgers} (a), the prediction error roughly increases with $N_s$, while the training time per step decreases. This suggests that more frequent updates of training points are beneficial for improving prediction accuracy, albeit at a modest increase in computational cost. For a compromise between accuracy and efficiency, we select $N_s=100$ as the default value.
For the clipping parameter $\gamma$ shown in Fig. \ref{Fig_Hyperparameter_Burgers} (b), the prediction error exhibits an approximately concave trend, with the lowest error occurring at about $\gamma=100$. This suggests that clipping is necessary to prevent excessive focus on a few points with very large residuals, which can lead to overfitting. However, overly aggressive clipping (i.e., too small $\gamma$)  will diminish the effectiveness of adaptive sampling, making it similar to random sampling. Therefore, we choose $\gamma=100$ as the default value. The training time per step remains relatively constant across different $\gamma$ values with only slight decrease with increasing $\gamma$.
For the fraction of points updated $p_u$ shown in Fig. \ref{Fig_Hyperparameter_Burgers} (c), the prediction error is very large when $p_u$ is too small (e.g., $p_u=0.05$), as it will get close to non-adaptive sampling. As $p_u$ increases, it almost remains nearly unchanged with an approximately increasing trend with $p_u$. For a detailed comparison, Fig. \ref{Fig_Hyperparameter_Burgers} (d) shows the raw and smoothed prediction error histories for three selected $p_u$. The raw error curves are quite noisy, making it difficult to discern clear trends. To enhance visual clarity, we apply a moving minimum filter with a window size of 1500 training steps to smooth the error histories. The smoothed curves reveal that $p_u=0.2$ and $p_u=0.6$ yield similar performance, while $p_u=1$ results in significantly higher errors, especially at the final stage of training. This indicates that updating all points at each update step is not optimal, as it may lead to excessive fluctuations in the training set, hindering convergence especially when the prediction error is already low. The training time per step remains nearly constant across different $p_u$ values. Based on these observations, we recommend to use $p_u \in [0.2, 0.6]$ and select $p_u=0.2$ as the default value.

\section{Point and weight distribution after training}\label{sec_distribution}
\begin{figure}[htbp]
	\centering	
	\includegraphics[ width=10cm, trim=1.0cm 0.2cm 2.0cm 1.2cm, clip=true]{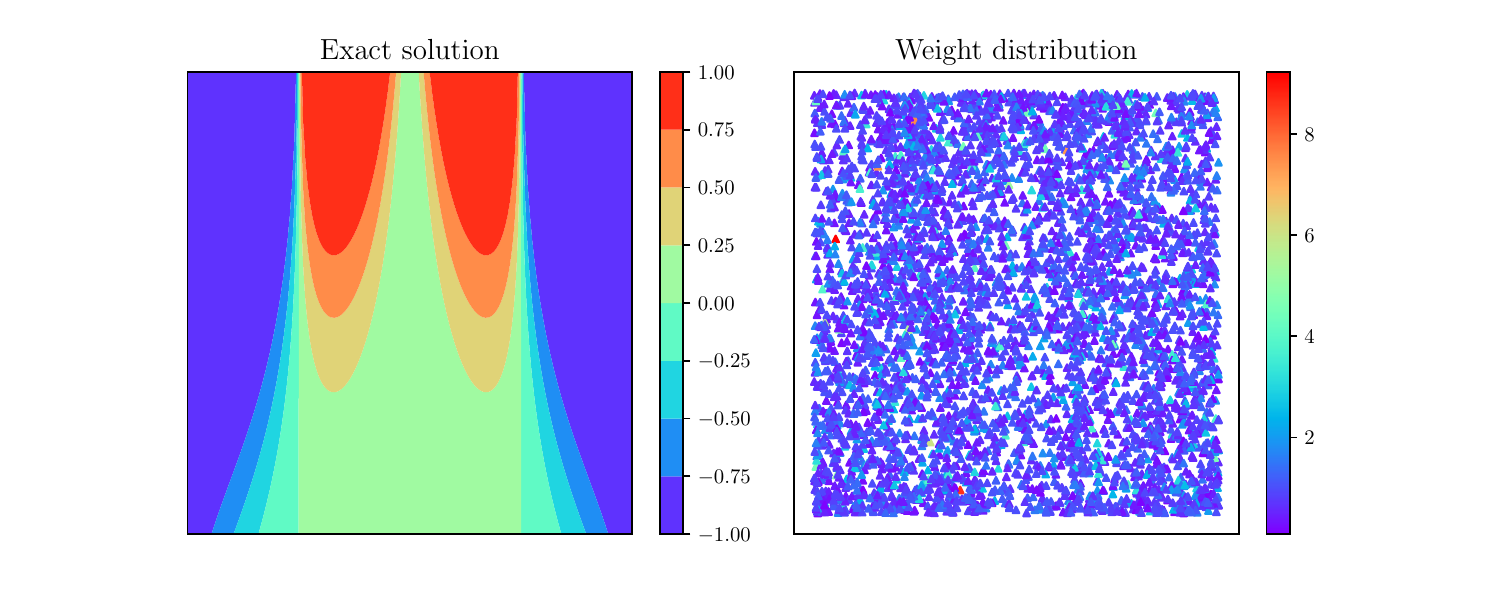}
	\caption{Allen-Cahn equation: (left)Exact solution and (right) weight distribution (color) over the residual points (scatter).
	}
	\label{Fig_distribution_AllenCahn}	
\end{figure}

\begin{figure}[htbp]
	\centering	
	\includegraphics[ width=10cm, trim=1.0cm 0.2cm 2.0cm 1.2cm, clip=true]{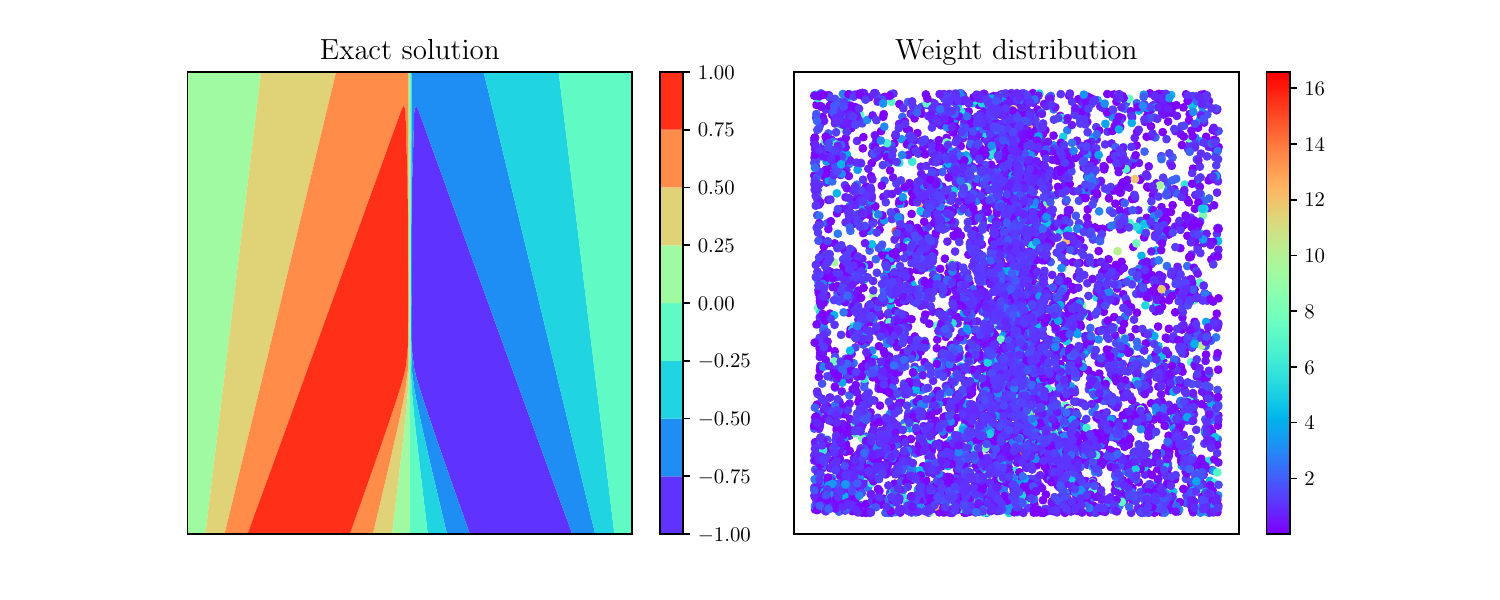}
	\caption{Burgers equation: (left)PINN prediction errors and exact solution and (right) weight distribution (color) over the residual points (scatter).
	}
	\label{Fig_distribution_Burgers}	
\end{figure}

\begin{figure}[htbp]
	\centering	
	\includegraphics[ width=10cm, trim=1.0cm 0.2cm 2.0cm 1.2cm, clip=true]{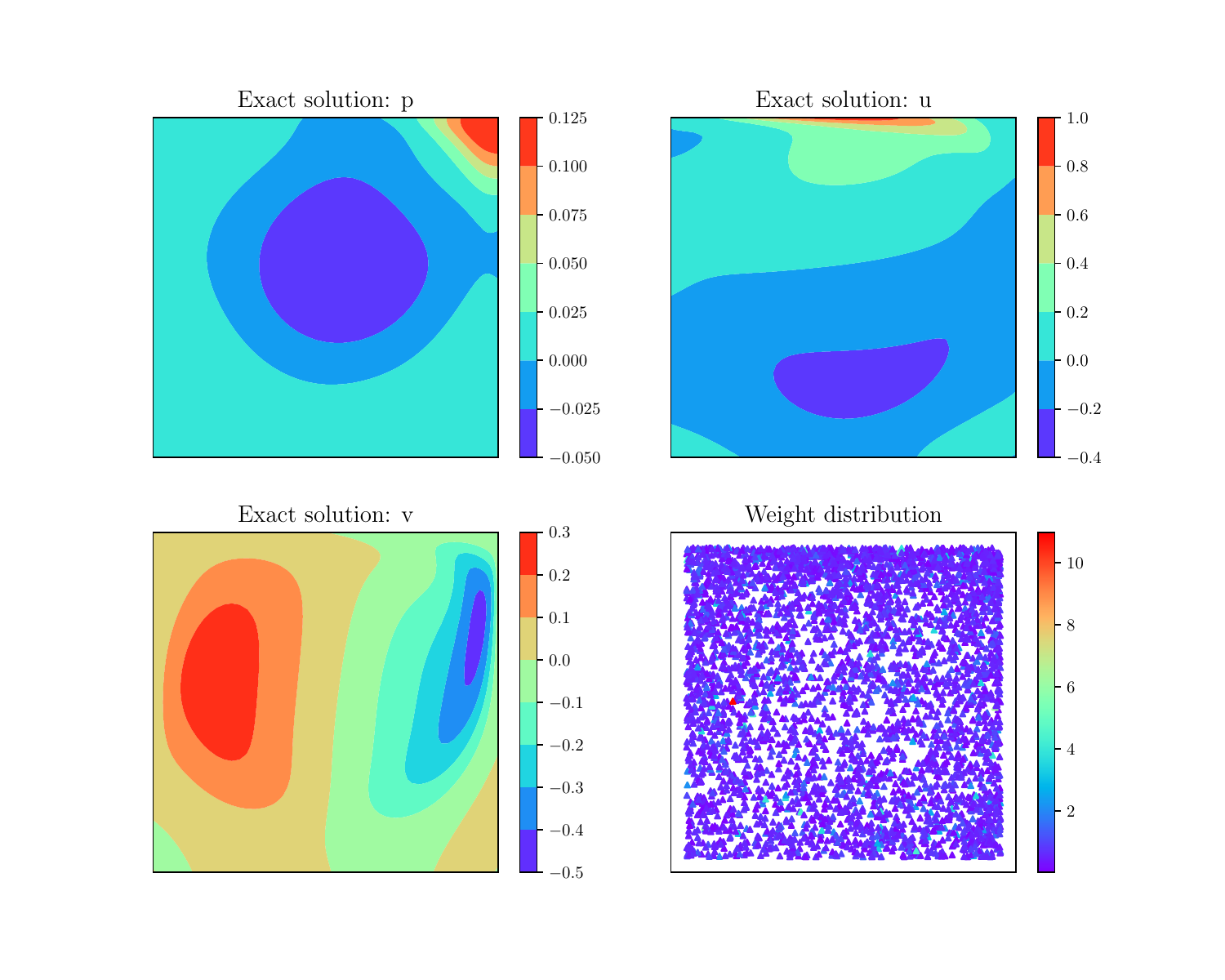}
	\caption{Lid-driven cavity flow: (left)PINN prediction errors and exact solution and (right) weight distribution (color) over the residual points (scatter).
	}
	\label{Fig_distribution_LidDriven}	
\end{figure}

The distributions of training points and weights after applying both adaptive sampling and adaptive weighting are illustrated in Figs. \ref{Fig_distribution_AllenCahn}, \ref{Fig_distribution_Burgers}, and \ref{Fig_distribution_LidDriven} for the Allen–Cahn equation, Burgers equation, and lid-driven cavity flow problem, respectively. The results reveal distinct patterns in relation to the exact solutions. For the Allen–Cahn and Burgers equations, the training points are concentrated in regions with large gradients, while for the lid-driven cavity flow, they are concentrated near the walls. In contrast, the weight distributions appear nearly uniform across the domains for all three problems.

%\section*{References}
\bibliographystyle{elsarticle-num}
\bibliography{bibliography}

\end{document}